\documentclass[sn-apa]{sn-jnl}

\jyear{2022}%

\theoremstyle{thmstyleone}%
\theoremstyle{thmstyletwo}%

\theoremstyle{thmstylethree}%

\begin{document}

\title[Triplet-loss Dilated Residual Network for High-Resolution Representation...]{A Triplet-loss Dilated Residual Network for High-Resolution Representation Learning in Image Retrieval}

\author[1]{\fnm{Saeideh} \sur{Yousefzadeh}}\email{sa.yousefzadehmaghani@mail.um.ac.ir}

\author*[1]{\fnm{Hamidreza} \sur{Pourreza}}\email{hpourreza@um.ac.ir}

\author[2]{~\fnm{Hamidreza} \sur{Mahyar}}\email{mahyarh@mcmaster.ca}

\affil[1]{\orgdiv{Machine Vision Lab}, \orgname{Ferdowsi University of Mashhad}, \country{Iran}}

\affil[2]{\orgdiv{Faculty of Engineering}, \orgname{McMaster University}, \country{Canada}}

\abstract{Content-based image retrieval is the process of retrieving a subset of images from an extensive image gallery based on visual contents, such as color, shape or spatial relations, and texture. In some applications, such as localization, image retrieval is employed as the initial step. In such cases, the accuracy of the top-retrieved images significantly affects the overall system accuracy. The current paper introduces a simple yet efficient image retrieval system with a fewer trainable parameters, which offers acceptable accuracy in top-retrieved images. The proposed method benefits from a dilated residual convolutional neural network with triplet loss. Experimental evaluations show that this model can extract richer information (i.e., high-resolution representations) by enlarging the receptive field, thus improving image retrieval accuracy without increasing the depth or complexity of the model. To enhance the extracted representations' robustness, the current research obtains candidate regions of interest from each feature map and applies Generalized-Mean pooling to the regions. As the choice of triplets in a triplet-based network affects the model training, we employ a triplet online mining method. We test the performance of the proposed method under various configurations on two of the challenging image-retrieval datasets, namely Revisited Paris6k (RPar) and UKBench. The experimental results show an accuracy of 94.54 and 80.23 (mean precision at rank 10) in the RPar medium and hard modes and 3.86 (recall at rank 4) in the UKBench dataset, respectively.}




\maketitle

\section{Introduction}\label{sec1}

\begin{figure*}[h!]
\begin{center}
\centering
\includegraphics[width=0.6\textwidth]{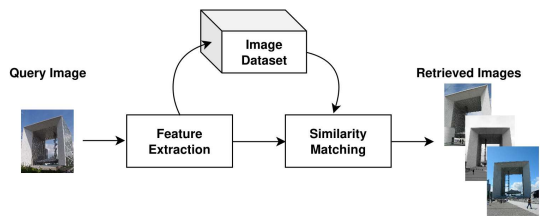}
\caption{Illustration of the image retrieval system. It aims to retrieve all images
that contain the same content as the query image from
a sizeable unordered collection of images. The basic principles of CBIR are feature extraction and similarity measurement.}\label{fig1}
\end{center}
\end{figure*}

An old research topic in computer vision is content-based image retrieval (CBIR), whose aim is to retrieve similar images in an extensive image gallery by analyzing their visual content (Figure~\ref{fig1}). The applications of CBIR include visual geo-localization \cite{bib1,bib2}, medical image search \cite{bib3}, person re-identification (Re-ID) \cite{bib4}, 3D reconstruction \cite{bib5}, remote sensing \cite{bib6}, shopping recommendations in online markets \cite{bib7}, and many others.

The primary methods of image retrieval rely on powerful hand-crafted features, such as SIFT\footnote{Scale Invariant Feature Transform} , and encoding methods, such as BoW\footnote{Bag Of Words}, VLAD\footnote{Vector Locally Aggregated Descriptor}, and Fisher Vector. Robust to scaling, translation, and rotation changes, SIFT and BoW can also represent images. VLAD has achieved significant results in capturing a compact image representation . Combining these methods with compression produces global descriptors that can be scaled to more extensive databases but at the expense of accuracy. These methods can also integrate with post-processing strategies, such as query expansion. Since 2012, with deep learning, much research has focused on convolutional neural network (CNN) models as image representation extractors for a content-based image search. Research has proved that superior semantic information is attained if CNN is utilized to extract image features \cite{bib8}. The usage of deep learning also reduces the semantic gap, a challenging issue in CBIR. The origin of this gap is the difference between the interpretation of images by humans and computers. Humans use high-level concepts to express visual content. In contrast, computers extract low-level features from image pixels. Deeper architectures assist in learning higher-level abstract features to reduce the semantic gap \cite{bib9}.

Several earlier methods apply CNN activation to image retrieval \cite{bib8,bib10,bib11}. Some of these methods use fully connected layer (FC) activation as a global image descriptor. However, research has shown that features built on fully connected layers are less ideal for image representation  \cite{bib12}. Each neuron in the FC layer is related to all previous neurons and has a global receptive field but is limited in terms of geometric invariance and spatial information \cite{bib13}. The features of the convolutional layers (usually the last layer) preserve further structural details useful for applications, such as retrieval \cite{bib14,bib15}.

The usage of the pooling layer after the convolutional layers improves the robustness of the convolutional features \cite{bib16}. Sum/average pooling and max pooling are two standard methods applied to convolutional feature maps, hybrid pooling and weighted average pooling are a few other suggested methods \cite{bib17}. Tolias et al. \cite{bib11} propose Regional Maximum Activations of Convolutions (R-MAC) that aggregates activation features of convolutions in different regions of the image using a multi-scale rigid grid with overlapping cells. The present work also utilizes a regional pooling mechanism and applies GeM pooling to the regions, such as the regional generalized-mean pooling proposed by Weiqing \cite{bib18}. The GeM layer is a trainable pooling layer that generalizes max and average pooling to preserve the information of each channel. Usage of the GeM pooling layer has been shown to boost retrieval performance \cite{bib19}.

\begin{figure}[t]
\centering
\includegraphics[width=119mm]{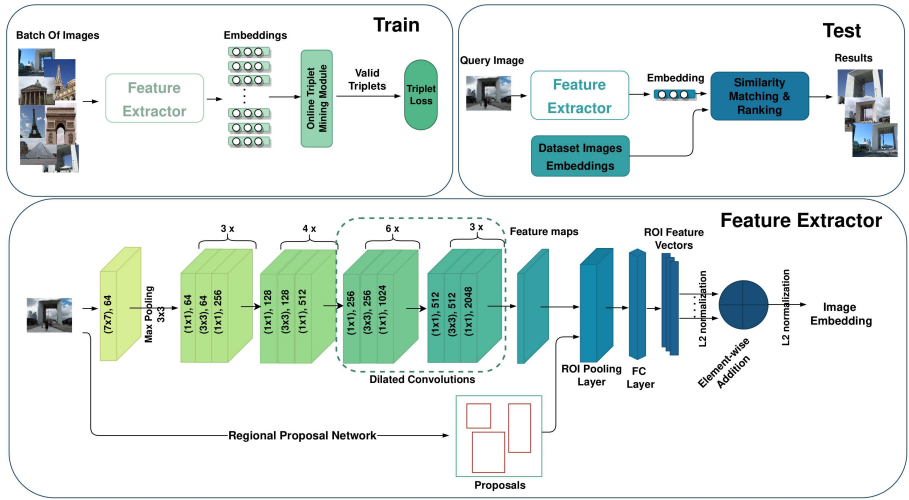}

\caption{The proposed architecture. A mini-batch of images is entered into the feature extractor in the training phase. We identify the valid triplets using the proposed triplet online mining module, and the triplet loss function calculates the corresponding error. In the test phase, after the extraction of query image features, similar images from the database are found by using the similarity criterion, and the results are ranked. The proposed feature extractor is based on the idea of dilated residual convolutional layers. The obtained feature maps and their regions of interest using the region proposal network (RPN) are passed through the Pooling and FC layers. After normalization, the features vectors are added together element-wise and normalized to get the image representation.}\label{fig2}
\end{figure}

The current work considers image retrieval as a deep metric learning (DML) problem. The architecture of the proposed method is depicted in Figure \ref{fig2}, which will be then explained in more details in Section \ref{sec3}. First, in DML, deep neural networks are utilized to embed images in an embedded metric space. Then the similarity between images can be measured using simple criteria, such as the Euclidean distance or the cosine distance. Siamese \cite{bib20} and triplet networks \cite{bib21,bib22} are typical architectures for metric learning. The most common loss function employed in DML is the triplet loss function that works with triplets (anchor, positive, negative) \cite{bib20}. The selection of triplets is essential for efficient training \cite{bib23}. Various applications have followed different triplet selection methods \cite{bib20,bib21,bib24,bib25}. Many image retrieval tasks employ the offline method for choosing triplets, an inefficient process by which triplets are randomly selected from the entire training dataset after each training epoch \cite{bib26}. In their online triplet mining technique, Schroff et al. pick valid triplets from each mini-batch in training, which results in easier convergence \cite{bib20}.

The online selection of triplets can face challenges when there are unrelated images and a few similar images of each class in the dataset. Besides these problems, the datasets used for localization also encounter the issue of changes in the appearance of images. Different seasons, altering weather conditions, the illumination at other times of the day, dynamic objects, such as pedestrians or cars passing by, and changes in viewpoints cause changes in image appearance. Image retrieval tasks that deal with images with such changes and challenges often use the offline method and randomly select triplets \cite{bib18,bib22,bib27}. Several studies have also employed 3D image information as a triplet mining solution for valid triplet selection \cite{bib19,bib28}. The 3D reconstruction process is computationally expensive and requires much memory for the map. The current paper presents a deep triplet-based network with an online triplet mining module. The proposed method has yielded promising results on datasets that meet the challenging conditions mentioned above.

As known, one necessity for any image retrieval system is to display the features of the images distinctly. Consequently, deep CNN must be designed to extract features from images that contain helpful and discriminative information. For this purpose, the current research uses the idea of dilated convolutional networks. As an image passes through the layers of the convolutional networks, its resolution gradually decreases so that the spatial structure of the scene is no longer recognizable by the small output feature map obtained at the end of the network. Furthermore, this reduces the accuracy problems where a detailed understanding of the scene is necessary (e.g., semantic segmentation) \cite{bib29}. Dilation can ease these problems and is an efficient technique for increasing the network receptive field without the loss of resolution. Researchers have studied dilated convolutional networks in different problems, such as semantic segmentation and image classification,  and achieved excellent results \cite{bib29,bib30,bib31}.

The current research has shown that dilation can extract richer information to improve image retrieval accuracy without increasing the depth or complexity of the model. Thus, tasks that deal with complex and challenging images and need an accurate understanding of the scene may use dilated networks.

One feature of the work performed is the model's simplicity and fluent training without added complexity and calculations. CNN-based image retrieval methods vary in complexity, growth capacity, and execution time. As known, in many applications, such as some methods introduced for position estimation and navigation and landmark recognition, retrieving similar images is an initial step followed by other subsequent processing steps \cite{bib32,bib33}. Moreover, the accuracy of the top retrieved images is critical. Therefore, designing a simple model that can accurately retrieve similar images without creating more complexity is necessary.

The main contributions of this paper are summarized as follows:
\begin{itemize}

  \item In image retrieval, the accuracy of the top-retrieved images is usually important because it is visually checked by a user (recommendation systems). Most times, such as localization, image retrieval is an initial step followed by subsequent processing steps. Therefore, the top retrieved images must be correct. The primary emphasis in this work is to offer a simple model with low training parameters and no added complexity that can provide acceptable accuracy in top-retrieved images.

  \item The present study uses a deep residual CNN with dilated convolution layers to capture high-level features with larger receptive fields and to produce high-resolution density maps without expanding network depth or complexity. The present study results indicate that residual dilated networks can be helpful for image analysis and retrieval tasks since the output from dilated convolution includes information with more details. This is especially advantageous when there are complex natural images, and an accurate understanding of the scene is essential.

  \item This paper presents a deep triplet-based network with an online triplet mining module. The online nature of the proposed method is more efficient because it provides more triplets for one mini-batch and does not require offline extraction calculations. The proposed approach has yielded promising results on datasets that meet the challenging conditions such as changes in appearance, viewpoints, the incomplete view of the object, the lack of sufficiently similar images, and unrelated images. Image retrieval tasks that deal with images with such changes and challenges often use the offline method and randomly select triplets or employ 3D image information for valid triplet selection that is computationally expensive and requires much memory for the map.

\end{itemize}

The remaining parts of the present paper are as follows. Section~\ref{sec2} presents related works. Section~\ref{sec3} details the network’s architecture, including dilated residual networks, Generalized-Mean (GeM) pooling, triplet loss function, the selection of training image pairs and the similarity measure. Section~\ref{sec4} presents the database images used, the performance evaluation measures are discussed, and covers the experiments and results. Finally, Section~\ref{sec5} concludes.

\section{Related Work}\label{sec2}
Recent developments have shown that CNNs are the right choice for obtaining image search representations \cite{bib8,bib34}. The resulting representations are compact, efficient for searching, and highly differentiated. Some image retrieval methods use the output of the last fully connected layer as global image descriptors. Recently, research has focused on features extracted from CNN's deep convolution layers. The training of CNNs is possible from scratch or by fine-tuning. In fine-tuning, a pre-trained network, which is not necessarily optimal for a study's work, can be retrained for an intended task, thus resulting in significant improvement in adaptability. Babenko improves the pre-trained models on ImageNet by fine-tuning them on a Landmark dataset \cite{bib35}. Other studies follow this concept by fine-tuning the pre-trained models for image retrieval or employing ranking loss, thus resulting in significant improvements \cite{bib1,bib19,bib28}. Gordo et al. propose a triplet network with a ranking loss to produce discriminative feature representation \cite{bib22}. Razavian utilizes the activations of convolutional layers followed by a global-pooling aggregation \cite{bib14}. Feature aggregation, a feature enhancement method, improves the discriminative ability and robustness of in-depth features. Babenko et al. suggest SPoC, in which a sum-pooling mechanism obtains compact global descriptors for image retrieval \cite{bib10}. Kalantidis et al. extend SPoC by allowing cross-dimensional weighting and the aggregation of neural codes \cite{bib17}. Tolias et al. Propose R-MAC, which aggregates the activation features of convolutions in several image regions into a compact feature vector of fixed length \cite{bib11}. \cite{bib19} propose a novel trainable Generalized-Mean (GeM) pooling layer that generalizes max and average pooling to preserve the information of each channel.

Research has indicated that usage of the GeM pooling layer boosts retrieval performance  \cite{bib19}. Consequently, the present research has included this pooling layer in its network architecture. Similar to \cite{bib18}, the current study has obtained candidate regions of interest from each image and applied GeM pooling to the areas.
By localizing the candidates of regions in images, the Region Proposal Network (RPN) enables deep models to learn regional features for particular objects \cite{bib22,bib36}. Similar to \cite{bib27}, the present research implements RPN with a fully convolutional network constructed on top of the convolutional layers and trains it with bounding boxes estimated for the Landmark images \cite{bib36}. This model provides the proposed regions at the cost of almost zero.

\textbf{Triplet networks and metric learning}. The present study considers image retrieval as a metric learning problem. The two-branch Siamese \cite{bib20,bib37} and triplet networks \cite{bib21,bib22} are typical architectures for metric learning which use matching and non-matching pairs of images to carry out the training. In these networks, matching pairs are similar and adjacent images that belong to the same class, and it is easy to consider two dissimilar images, taken far from each other and rarely belonging to the same class, as non-matching pairs. Many computer vision tasks successfully use triplet loss, such as image classification \cite{bib24}, face recognition \cite{bib20}, and metric learning \cite{bib38} whose usage of triplet loss is widespread. The performance of triplet loss is much affected by how triplets are selected \cite{bib23}. It can be said that, if triplets are randomly chosen, convergence becomes difficult or, if the hardest triplets are selected, this often leads to a bad local optimum \cite{bib39,bib40,bib41}.

\sloppy Different applications have used various triplet selection methods \cite{bib20,bib21,bib24,bib25}. Many image retrieval tasks randomly select triplets from the entire training dataset after each training epoch \cite{bib18,bib26,bib27,bib38}; known as the offline method, this approach is inefficient. In contrast, triplets can be extracted from each mini-batch during the training process, a technique known as online triplet mining suggested in \cite{bib20}. Online techniques render the training process easier to converge. The research \cite{bib25} selects triplets (anchor, positive, negative) using the top k triplets in each mini-batch based on the margin, $distance(anchor,positive)-distance(anchor,negative)$, where $distance$ is the squared Euclidean distance. \cite{bib21} chooses only hard triplets, i.e., $distance(anchor,negative)<distance(anchor,positive)$, while \cite{bib20} picks semi-hard triplets, i.e., $distance(anchor,positive)+\alpha<distance(anchor,negative)$, where $\alpha$ is a positive scalar. The Simo-Serra work employs hard positive examples, in which samples are extracted with the help of three-dimensional reconstruction \cite{bib42}. Hard positive pairs must be chosen with great care to avoid overfitting \cite{bib41,bib43}.

The challenges of choosing triplets online can be images with changes in appearance, unrelated images, images with a low and incomplete view of the object or landmark, and a lack of sufficiently similar images from each class. Most image retrieval tasks that work on such images, e.g., landmark images, use the offline method \cite{bib18,bib22,bib27}. Several studies have also employed 3D image information as a triplet mining solution for selecting valid triplets \cite{bib19,bib28}. Nevertheless, the 3D reconstruction process is computationally expensive, and its need for much memory for the map often prevents this process from being implemented on resource-constrained platforms, such as aerial robots.
Therefore, it is essential to adopt a simple triplet selection method to extract meaningful and discriminative triplets from mini-batches online and achieve promising results on datasets despite the challenges mentioned above.

\textbf{Dilated convolution layers}. Dilated convolution layers were first introduced in 2016 \cite{bib29}. Because of performing a sparse sampling of input feature maps to increase the spatial scale of output feature maps, they differ from regular convolution filters. Dilated convolutional layers have been applied to segmentation tasks and have achieved significant improvements in accuracy without introducing additional parameters or computational costs \cite{bib29}. The \cite{bib29} study shows that dilated convolution is advantageous for dense prediction. Their model uses the general multi-scale context information of the dilated convolution system without losing its resolution. The usage of dilated convolution obtains more pertinent information with enlarging the receptive field. Chen proposes the DeepLab system that controls the resolution of feature maps computed within CNNs by dilated convolution \cite{bib31}. This system has made significant progress in image recognition. The \cite{bib30} research shows that dilated residual networks (DRN) improve image classification performance. In particular, without increasing the depth or complexity of the model, DRNs are more accurate in ImageNet classification than their non-dilated counterparts \cite{bib30}. DRN has the same number of parameters and layers as the original Residual Network (ResNet). The results of the present study indicate that residual dilated networks can be helpful for image analysis and retrieval tasks since the output from dilated convolution includes information with more details. This is especially advantageous when there are complex natural images and accurate understanding of the scene is essential.

\section{Network Architecture and Image Representation}\label{sec3}

The purpose of the CBIR is to retrieve images from the dataset that contains the most visual content that is similar to the query image. The basic principles of CBIR are feature representation and similarity measurement. Therefore, the performance of the CBIR method strongly depends on how the features of the image are represented. The fundamental idea of the proposed model is to design a deep CNN for extracting high-level features with larger receptive fields and generating high-quality density maps without expanding network complexity (Figure~\ref{fig2}). Research has shown that residual networks reduce the semantic gap, especially when the network is trained with the ranking loss ResNets can learn a more invariant representation of the regions and mitigate the background effect. \cite{bib27}. The current study does not utilize the original ResNet. Instead, a dilated residual network (DRN) has been employed that significantly expands the network's vision and can also combine further contextual information with a lower computational cost. Section~\ref{subsec3.1} describes the DRN.

The proposed model discards fully connected layers of DRN and uses a Region of Interest (ROI) pooling layer that follows the last convolutional layer and performs GeM pooling on the regions. Section~\ref{subsec3.2} introduces the GeM layer. The usage of the pooling layer after the final convolutional layer improves robustness. The present work utilizes the Region Proposal Network (RPN) to localize candidate regions and center suggested regions on the objects of interest. After the pooling, a fully connected (FC) layer with learnable weights is placed. After crossing the FC layer, the pooled features of regions are independently L2-normalized (L2N). They are then sum-aggregated and L2-normalized once again and produce a compact vector. The sum-aggregation of different regions and the L2-normalization are differentiable.

The present research employs the idea of triplet networks that receive triplets of inputs and combine three input streams with a triplet loss. Research has shown that triplet loss is more effective for ranking problems. Section~\ref{subsec3.3} introduces this loss function, while Section~\ref{subsec3.4} provides additional details on how to choose triplets and the similarity measure is discussed in Section~\ref{subsec3.5}.
Most works based on triplet networks comprise three separate streams, each of which is a copy of the feature extractor, with parameters and weights shared among them. Utilizing three copies of the model with shared parameters is a satisfactory optimization approach. However, it is inefficient because of computing and memory constraints \cite{bib39} due to training with large images and three streams simultaneously. With a more extensive network, such as ResNet101, there is not enough memory to process even one triplet, since the condition becomes even more complex \cite{bib27}. The current paper only uses only one stream and selects triplets online from within the mini-batches. This allows for training the model by using deep architectures and without reducing the size of the training images.

\begin{figure}[t]
\centering
\includegraphics[width=0.4\textwidth]{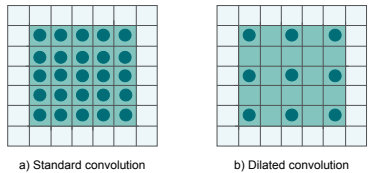}
\caption{Dilated convolution supports larger receptive fields compared to standard convolution. The filter size in (b) is only 3*3, but the receptive field of (b) is the same as (a).}\label{fig3}
\end{figure}

\subsection{Dilated Residual Networks}\label{subsec3.1}

Convolution networks gradually reduce image resolution and acquire compact feature maps in which the spatial structure of the scene is less recognizable. This can limit image classification accuracy and make it difficult to transfer the model to other applications that require accurate scene perception \cite{bib29,bib30}. Based on stationary wavelet transform idea, the dilation mechanism solves these problems. It is an effective technique for expanding the network receptive field without a loss of resolution or coverage. Dilated convolutions increase the spatial scale of output feature maps via the holes in their receptive fields \cite{bib29} (Figure~\ref{fig3}). Eq.~\ref{eq1} and~\ref{eq2} are a standard and dilated convolution, respectively. The expression, $s+dt=p$, signifies that some points during convolution will be skipped. In these equations, $F$ is the filter associated with layer $G$ and the domain of $p$ is the feature map in $G$.

\begin{equation}
(G*F)(p)=\sum\limits_{s+t=p}G(s)F(t) \label{eq1}
\end{equation}

\begin{equation}
(G*F)(p)=\sum\limits_{s+dt=p}G(s)F(t) \label{eq2}
\end{equation}

In 2017, Yu et al. proposed dilated residual networks (DRN) that improve not only semantic segmentation but also image classification without increasing depth or complexity \cite{bib30}. Recently, residual networks (ResNets) in various computer vision tasks have achieved success, proving that ResNet has a better image representation capacity than other deep architectures. ResNets can extract conceptual features and more details to display image features. DRN's result from applied dilated convolutions in the residual blocks and have both the properties of the residual network and the advantages of the dilation mechanism.
The number of layers and parameters of DRN is the same as the original ResNet. The fundamental difference is that the original ResNet samples the input image with a factor of 32 and the DRN with a factor of 8.  In the original ResNet, the final two groups of convolutional layers use a 3$*$3 standard convolution (dilation=1) and the output feature map is the size of 7$*$7. In DRN, Group4 uses a dilation of 2 while Group5 continues with dilation of 2 for the first convolution and 4 for the remaining convolution.

\begin{figure}[t]
\centering
\includegraphics[width=0.6\textwidth]{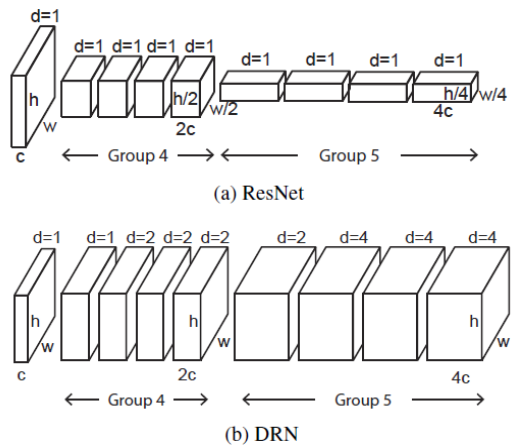}
\caption{Convert ResNet to DRN. In the original ResNet, the final two groups of convolutional layers use a 3*3 standard convolution (dilation=1), and the output feature map is size 7*7. In DRN, Group4 uses a dilation of 2, while Group5 uses a dilation of 2 for the first convolution and 4 for the remaining convolution. The output of Group5 in DRN is 28×28, which is much larger than that of the original ResNet (Figure from \cite{bib30}).}\label{fig4}
\end{figure}

\begin{table}[t]
\begin{center}
\begin{minipage}{165pt}
\centering
\caption{Different models of DRN and ResNet and the number of their parameters}\label{tab1}
\begin{tabular*}{13pc}{@{\extracolsep{\fill}}@{}cc@{}}
\toprule
Model & Parameters \\
\midrule
ResNet-18   & 11.7 M  \\
DRN-A-18    & 11.7 M  \\
DRN-B-26    & 21.1 M  \\
DRN-C-26    & 21.1 M  \\
\midrule
ResNet-34    & 21.8 M  \\
DRN-A-34    & 21.8 M  \\
DRN-C-42    & 31.2 M  \\
\midrule
ResNet-50    & 25.6 M  \\
DRN-A-50    & 25.6 M  \\
\midrule
ResNet-101  & 44.5 M  \\
\botrule
\end{tabular*}
\end{minipage}
\end{center}
\end{table}
\space

The output of Group5 in DRN is 28$*$28, which is much larger than that of the original ResNet (Figure~\ref{fig4}).
Table~\ref{tab1} presents different models of DRN and the number of their parameters. The present research uses DRN-A-50 in its experiments. Despite having the same depth and capacity, each DRN-A outperforms its respective ResNet model. The results show that DRN networks can be used in image analysis tasks that involve complex natural images, especially when it is necessary to understand scenes accurately \cite{bib30}.

\subsection{Generalized-Mean Pooling Layer (GeM)}\label{subsec3.2}

Image retrieval methods often use average and max pooling as the primary aggregation mechanism. However, these pooling layers pose drawbacks. Max-pooling only computes the maximum value of each feature map, so it may lose information on other activations. Sum-pooling or average pooling averages all activations of each feature map, which will also undermine the discrimination of the final representation \cite{bib16,bib44}. Radenovic et al. suggest the generalized mean (GeM), a novel trainable pooling that generalizes max and average pooling \cite{bib19}. GeM has had a significant performance compared to non-trainable standard pooling layers. The corresponding GeM descriptor is given by:
\begin{align}
f^{(GeM)} &=  \left[ f_1^{(GeM)} f_2^{(GeM)} f_m^{(GeM)}…f_M^{(GeM)}  \right]^T  \nonumber \\
f^{(GeM)}_m &= \left( \frac{1}{\lvert X_m\lvert} \sum\limits_{x \epsilon X_m} x^{P_m} \right)^\frac{1}{P_m} \label{eq3}
\end{align}
In Eq.~\ref{eq3}, max-pooling and sum-pooling are the specific cases of GeM pooling. When $p\longrightarrow+\infty$, GeM pooling turns into max-pooling and, if p equals 1, GeM pooling turns into average-pooling. Let $X$ is a 3D tensor of $W$ × $H$ × $M$ dimensions, where $M$ is the number of feature maps in the last layer. In this equation, $P_m$ is a particular pooling parameter for each feature map, $X_m$, or a common parameter, p, can be used for all feature maps. Finally, the feature vector comprises a single value per feature map. The dimension of this vector is equal to $M$ which is 256, 512, or 2048, on many popular networks, making it a compressed image representation. With the usage of ResNet-101, $M$ is 2048, and so each descriptor is a 2048-D vector. The current paper employs GeM pooling in a regional pooling mechanism, such as the regional generalized-mean pooling used in \cite{bib18}.

\subsection{Triplet Loss Function}\label{subsec3.3}

\begin{figure}[t]
\centering
\includegraphics[width=0.6\textwidth]{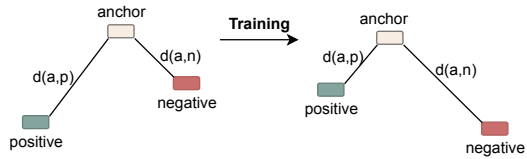}
\caption{Illustration of the triplet loss. The triplet loss function minimizes the anchor distance to the positive image and maximizes the anchor distance to the negative image.}\label{fig5}
\end{figure}

The triplet loss is the most common loss function used in ranking problems. This function works with the triplets of samples: anchor, positive (similar to the anchor sample), and negative (not similar to the anchor sample). This loss function minimizes the distance from the anchor to the positive image and maximizes it from the anchor to the negative image (Figure~\ref{fig5}).
The distance between the representations of samples can be calculated (for example, the Euclidean distance) and the model can be optimized with this loss function. Thus, this model learns similar representations for samples defined as similar and distant representations for those not defined as similar. The triplet loss function is::
\begin{equation}
tripletloss(a,p,n)=max(distance(a,p)-distance(a,n)+margin,0)\label{eq4}
\end{equation}
where $margin$ is the max-margin and $distance$ is the squared Euclidean distance (the negative dot product may also be used instead), and $a$, $p$, and $n$ are representations of the anchor, positive, and negative image, respectively. Triplet loss optimizes the model so that the distance from the representations of an anchor to a negative image is over one margin relative to the desired distance to a positive image. This loss has achieved significant success in instance-wise recognition tasks \cite{bib45,bib46}.

\subsection{Selection of Training Image Pairs}\label{subsec3.4}

As typical architectures for metric learning, triplet networks require the input of 'triplets' (anchor, positive, negative) for training with triplet loss. Choosing triplets is essential for efficient training. The usual way to generate triplets is to find them at the beginning of each epoch. This approach known as the offline method. This method calculates all the representations of the training set and then selects only the hard or semi-hard triplets. Based on its results, the offline method has not been effective, because producing triplets requires conducting a complete review of the training set. This method also entails regular updating of the extracted triplets offline. The method of the online mining of triplets proposed in \cite{bib20} calculates the embedding of images in mini-batches (size $B$), from which a maximum of $B^3$ triplets can be extracted, although not all of these triplets are valid. The online method is more efficient because it provides more triplets for one mini-batch and does not require offline extraction calculations.

Most image retrieval tasks that work on datasets containing natural images, such as landmarks, use the offline method. These datasets present a large intra-class variability, with various views and a non-negligible number of unrelated images. Some works use a 3D reconstruction process to find appropriate triplets which are complex computational activities.

The current research employs an online triplet extraction method that yields promising results on datasets with these challenges. After the representation of the mini-batch images are obtained, the distance matrix of the image pairs is constructed based on the Euclidean distance. Then, 2D masks detect the valid pairs, namely (anchor, positive) and (anchor, negative). An (anchor, positive) pair is valid if the pair is separate and has the same label, while an (anchor, negative) pair is valid if the pair  is separate and does not have the same label.
For fast convergence of triplet-based networks, only beneficial triplets must be constructed, i.e., triplets with a positive loss. Depending on the distance between the triplet samples, there can be three triplets in the loss calculation: easy, hard, and semi-hard. Each of these definitions depends on the location of the negative sample relative to the anchor and positive sample. As a result, these three categories can be extended to easy negative, hard negative, and semi-hard negative.

Compared with the offline triplet-selection strategy, the online triplet mining method increases the potency of training by exploiting all the valid triplets online within a training mini-batch. Because the hard samples of each training batch are fully used to compute loss, the training process is easier to converge \cite{bib26}.
For the negatives, the present study chooses the hardest or the images that most closely resemble the anchor but belong to a class other than the anchor class. Many works have discussed the benefit of hard negative mining in creating triplets that deliver appropriate gradients and help triplet networks converge quickly \cite{bib20,bib43}. Nevertheless, this is not the case with hard positive examples. The \cite{bib43} study empirically shows that hard positive mining is not commonly appropriate for all datasets. The considerable intra-class variations require accurate sampling of positive pairs. If the model is forced to learn hard positive, this may lead to overfitting \cite{bib43}. In practice, the current work has also found that utilizing all the positive images speeds up the convergence. For this reason, instead of picking the hardest positive, the present research uses all (anchor, positive) pairs in a mini-batch while still selecting the hard negatives.

\subsection{Similarity Measure}\label{subsec3.5}

Different distance measures can be applied to feature vectors to compute the similarity among the query and images in the archive, e.g., the cosine distance or Euclidean distance \cite{bib47}. In the present work, these two criteria were tested and the results were better with the cosine similarity criterion. The results in the present work are obtained with the cosine metric which is:
\begin{equation}
cosine\_distance (A,B)=  1-cosine\_similarity (A,B) =1-\frac{(A.B)}{\|A\|.\|B\|} \label{eq5}
\end{equation}

In Eq.~\ref{eq5}, $A$ and $B$ are two n-dimensional vectors representing two images in the data space.

\section{Experimental Evaluation}\label{sec4}

\subsection{Dataset}\label{subsec4.1}

The current study performs experiments on two standard datasets, namely the Revisited Paris (RPar) \cite{bib48} and UKBench (UKB) \cite{bib49} datasets. Recently, the original Paris dataset has been revisited to correct annotation mistakes, add new query images, and introduce new evaluation protocols \cite{bib48}. The created dataset is referred to as RPar. This dataset is one of the biggest and most popular datasets featuring a great deal of in-class diversity, much noise, and unrelated images that complicate image retrieval tasks. The RPar dataset contains 6,412 images collected from Flickr\footnote{https://www.flickr.com/} by a search for 11 specific landmarks in Paris. It has 70 challenging queries and three new evaluation protocols of varying difficulty (easy, medium, and hard). The present paper reports the results of the medium and hard setups.
The UKB dataset comprises 10,200 images from 2,550 different groups. Each group has four images of a unique scene or object from various viewpoints, under several lighting conditions, and with different scaling. The images are from diverse categories, such as animals, plants, and household objects.

\subsection{Setting}\label{subsec4.2}

For the RPar dataset, all the images for each query image are ranked according to the cosine distance. Then, the current study selects the top K images from the ranked list as results, and reports the mean precision at K (mP$@$K) (K is usually 1, 5, 10) over the 70 queries \cite{bib47}. In image retrieval, usually the accuracy of the top-retrieved images is important because it is visually checked by a user (recommendation systems) and most times, such as in localization, image retrieval is an initial step followed by subsequent processing steps \cite{bib32,bib33}. Therefore, it is important that the top retrieved images are correct.
For UKB, the current research follows the standard evaluation protocol. For each group, one image is selected as the query and the goal is to retrieve the four nearest images to the query from the entire collection. Then, the recall is computed at four (recall$@$4) which is a number between 0 and 4.

\subsection{Evaluation Result and Discussion}\label{subsec4.3}

In its experiments, the present study implements the networks based on the PyTorch framework \cite{bib54}. First, we employ the Region Proposal Network (RPN) to localize regions of interest in images to produce candidate regions. The present research implements RPN with a fully convolutional network constructed on top of the convolutional layers and trains it with bounding boxes estimated for the Landmark images \cite{bib27}. For the convolutional part of our network, we consider a pre-trained DRN-A-50 without fully connected layers. Then, we aggregate features from different regions to produce a global image representation. The proposed model uses a Region of Interest (ROI) pooling layer that follows the last convolutional layer and performs GeM pooling on the regions. After the pooling, a fully connected (FC) layer with learnable weights is placed. After crossing the FC layer, the pooled region features are independently l2-normalized. They are then sum-aggregated and L2-normalized once again and produce a compact vector.

\begin{figure}[t]
\centering
\includegraphics[width=0.5\textwidth]{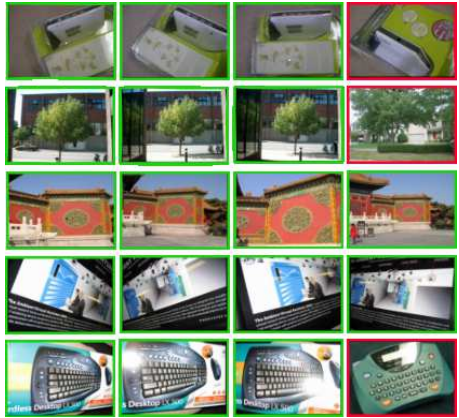}
\caption{Retrieval examples of UKBench (top-four matches). In each row, the first image from the left is used as query (The green and red boxes indicate the correct and incorrect retrieved image, respectively).}\label{fig6}
\end{figure}

For training, we employ the idea of triplet networks with a triplet loss. After the embedding of the mini-batch images is obtained, the distance matrix of the image pairs is constructed based on the Euclidean distance. Then, 2D masks detect the valid pairs, each image in the mini-batch is considered as an anchor and for which, among others in the mini-batch, the hardest negative and all available positives are calculated. Thus, triplets are formed by combining the hardest negative image with all positive image pairs in a mini-batch. To train the network, a stochastic gradient descent (SGD) optimizer is used with a momentum of $0.9$, a learning rate of $1e-4$, a weight decay of $5e-5$, and a batch size of 55. The margin of the triplet loss of 0.7 is empirically set. The training is performed for 20 epochs and all training images are resized to a maximum dimensionality of 240$*$240. During the training, the current research performs experiments on the central 228$*$228 crop of images. At test time, five crops with a 228$*$228 dimensionality are selected from each image, the center crop, and four corner crops. To extract their features, the crops are given to the model, and the average of the corresponding outputs is considered as the image representation. A dot-product is used to calculate the similarity between the query and images from the dataset and the results are ranked.

\begin{figure}[b]
\centering
\includegraphics[width=1\textwidth]{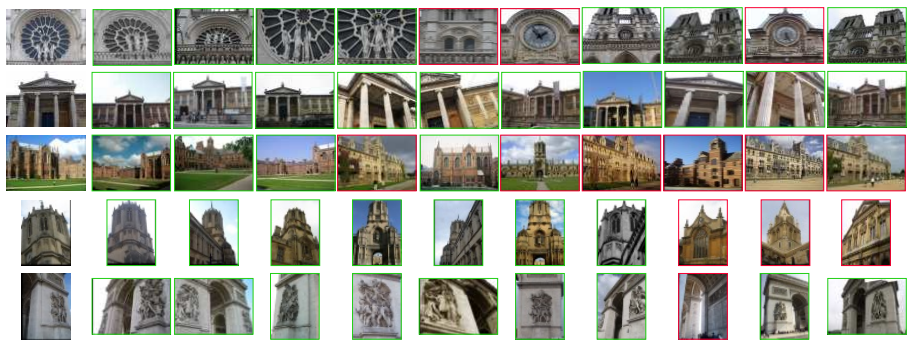}
\caption{Retrieval examples of RPar (top-ten matches). In each row, the first image from the left is used as query(The green and red boxes indicate the correct and incorrect retrieved image, respectively).}\label{fig7}
\end{figure}

The present work initializes the parameters of the networks by the corresponding network weights pre-trained on ImageNet. As usual, in the RPar dataset, only the query area of interest is used and, for the UKB dataset, the whole query image is employed. Figures~\ref{fig6} and~\ref{fig7} provide examples from retrieval images.

\begin{table}[t]
\begin{center}
\begin{minipage}{300pt}
\centering
\caption{Results of proposed method on RPar. Using mean precision at rank 10 (mP@10)}\label{tab2}
\begin{tabular*}{25pc}{@{\extracolsep{\fill}}@{}|lll|lll@{}}
\toprule%
\multicolumn{3}{c|}{medium}                                     & \multicolumn{3}{c}{hard}                                        \\ \hline
\multicolumn{1}{l}{mp@1} & \multicolumn{1}{l}{mp@5}   & mp@10 & \multicolumn{1}{l}{mp@1}   & \multicolumn{1}{l}{mp@5}  & mp@10 \\ \hline
\multicolumn{1}{l}{100}  & \multicolumn{1}{l}{96.114} & 94.54 & \multicolumn{1}{l}{95.714} & \multicolumn{1}{l}{85.12} & 80.23 \\
\botrule
\end{tabular*}
\end{minipage}
\end{center}
\end{table}

\textbf{Comparison with state-of-the-art.} For the RPar dataset, the performance is measured using mP$@$K (K is 1, 5, and 10) over the 70 queries presented in Table~\ref{tab2}. In Tables~\ref{tab3} and~\ref{tab4}, the current paper thoroughly compares the proposed framework with state-of-the-art methods in the image retrieval task. Performance is measured using mean precision at rank 10 (mP$@$10) in the Medium and Hard protocols for RPar, and (recall$@$4) for UKB. From the results reported, the following observations are noted:

In Table~\ref{tab4}, a comparison of the RPar results with other methods shows that the proposed method works better than the method presented in \cite{bib19}, based on the Alexnet architecture, and in \cite{bib28}, both in the Medium and Hard protocols. Compared to the ResNet-based method presented in \cite{bib19} and that introduced in \cite{bib27}, with a slight difference, the present study’s method shows an acceptable performance.

\begin{table}[h!]
\begin{center}
\begin{minipage}{190pt}
\centering
\caption{Results on UKB. Using recall at rank 4 (recall@4)}\label{tab3}
\begin{tabular}{@{}lc@{}}
\toprule%
Method          &  recal@4 \\ \hline
SPoC \cite{bib10}     & 3.65    \\ hline

Neural Codes \cite{bib35}            & 3.55    \\

Gordo et al. \cite{bib27}             & 3.84    \\

Multi-layer CNN \cite{bib52}         & 3.68    \\

\cite{bib53}             & 3.85    \\

R-mac \cite{bib11}                   & 3.74    \\

Proposed Method                & 3.86 \\
\botrule
\end{tabular}
\end{minipage}
\end{center}
\end{table}

\begin{table}[h]
\begin{center}
\begin{minipage}{280pt}
\centering
\caption{Comparison results of state-of-the-art methods on RPar. Using mean precision at rank 10 (mP@10) in the Medium and Hard protocols}\label{tab4}
\begin{tabular*}{23pc}{@{\extracolsep{\fill}}@{}lc c@{}}
\toprule%
method & medium & hard  \\ \hline
AlexNet-GeM \cite{bib19}     & 91.6   & 67.6  \\
ResNet101-GeM \cite{bib19}   & 98.1   & 89.1  \\
ResNet101-R-MAC \cite{bib27} & 96.9   & 86.1\\
AlexNet-MAC \cite{bib28}     & 92.9   & 69.3  \\
Proposed Method              & 94.54  & 80.23 \\
\botrule
\end{tabular*}
\end{minipage}
\end{center}
\end{table}

\begin{table}[h]
\begin{center}
\begin{minipage}{240pt}
\centering
\caption{Comparison of Accuracy density criterion of methods on RPar dataset}\label{tab5}
\begin{tabular}{@{}lc c@{}}
\toprule%
method                       & medium   & hard     \\ \hline
ResNet101-GeM \cite{bib19}   & 2.308e-6 & 2.096e-6 \\
ResNet101-R-MAC \cite{bib27} & 2.075e-6 & 1.843e-6 \\
Proposed Method        & 3.412e-6 & 2.895e-6 \\
\botrule
\end{tabular}
\end{minipage}
\end{center}
\end{table}

\begin{table}[t]
\begin{center}
\begin{minipage}{190pt}
\centering
\caption{Comparison of accuracy (mp@10) gained on PRar with different scenarios of triplet sampling methods}\label{tab6}
\begin{tabular*}{15pc}{@{\extracolsep{\fill}}@{}lc c@{}}
\toprule%
                                                                               & medium & hard  \\ \hline
\begin{tabular}[c]{@{}l@{}}Easiest-negative\\ Mean-positive\end{tabular}       & 92.63  & 78.84 \\
\midrule

\begin{tabular}[c]{@{}l@{}}Hardest-negative\\ Hardest-positive\end{tabular} & 90.34  & 77.41 \\
\midrule
\begin{tabular}[c]{@{}l@{}}Hardest-negative\\ Easiest-positive\end{tabular} & 91.68  & 78.21 \\
\midrule
\begin{tabular}[c]{@{}l@{}}Hardest-negative\\ Mean-positive\end{tabular}       & 94.54  & 80.23 \\
\botrule
\end{tabular*}
\end{minipage}
\end{center}
\end{table}

The primary emphasis in the current article is to offer a simple model with low training parameters and no added complexity that can provide acceptable accuracy in top-retrieved images. To prove the simplicity of the proposed model, the accuracy density criterion is employed to calculate the simplicity \cite{bib50,bib51}. Table~\ref{tab5} presents the results. The accuracy density criterion is obtained by dividing the model accuracy by the number of network parameters used. The higher the number obtained, the more desirable it is, and so the network achieves higher accuracy with fewer parameters \cite{bib50}. Although the proposed method has a slightly lower accuracy than the other two, the results show it is more straightforward and much less complex.

The results obtained on the UKB in Table~\ref{tab3} indicate that the proposed method outperforms the others mentioned in the table in retrieving four similar images related to each query image. In the absence of sufficiently similar images from each class, this database can be a good benchmark for measuring the proposed architectural strength. It should be noted that the proposed method has achieved satisfactory accuracy, despite its simplicity and the usage of the online selection of triples in the triplet network.

\textbf{Comparison of different scenarios of triplet sampling methods.} For triplet networks to converge faster, they must produce only useful triplets, that is, triplets with a positive loss. Various works have utilized different triplet exploration methods. The usage of the hardest negative image is familiar and improves performance, but choosing the hardest positive image does not work well for many datasets and leads to overfitting.
To construct triplets, the current work obtains the representation of a mini-batch of images that has entered, each image of which is then considered as an anchor and for which, among others in the mini-batch, the hardest negative and all available positives are calculated. Thus triplets are formed by combining the hardest negative image with all positive image pairs in a mini-batch.
To compare the results of its experiments, the current research also considers several proposed methods, such as the easiest positive, the easiest negative, the hardest positive, etc. The comparison shows that a continuous rise in performance is achieved by the use of increasingly diverse positive samples, which produce a greater diversity of viewpoints, and negative samples, with content closer to the anchor (Table~\ref{tab6}).

\textbf{Details of time cost.} The effectiveness of a CBIR system depends on the accuracy and the retrieval time in which it can produce the results. Retrieval time is a crucial attribute of CBIR techniques, especially in evaluating real-time applications. We can discuss it in terms of feature extraction time and total search time. In this work, with the effective image representation, we establish a search method for the CBIR system that explores the feature vectors via a similarity measure. The proposed CBIR is based on dilated residual networks that represented images with better efficiency because despite having the same depth and capacity, each DRN outperforms its respective ResNet model. The details of the time cost (feature extraction time and search time) of the proposed method on a single GPU (GeForce RTX 3090) are presented in Table~\ref{tab7}.

\begin{table}[t]
\begin{center}
\begin{minipage}{190 pt}
\centering
\caption{Details of time cost (second) of proposed method on RPar and UKB datasets}\label{tab7}
\begin{tabular}{@{}lc c@{}}
\toprule%
         & Feature extraction time   & Search time     \\ \hline
RParis    & 0.093 & 1.05 \\
UKBench  & 0.078 & 1.8 \\
\botrule
\end{tabular}
\end{minipage}
\end{center}
\end{table}

The main limitation of some image retrieval methods is that offline learning and training are extremely time-consuming. The Landmarks dataset presents a non-negligible amount of unrelated images. Some works preprocess the Landmarks dataset and clean the dataset to get the characteristics that need for training their models \cite{bib22,bib27}. This heavy cleaning procedure is performed at training time and is considered a major limitation of these algorithms. In \cite{bib19,bib28} is used a 3D reconstruction process to find appropriate triplets which takes higher training time and sometimes it goes memory is insufficient because of the complex computational activities. Especially for large datasets, the cost of construction of a 3d map and computational cost is large.

So the focus of the current work would be is to offer a simple and efficient model that despite having a low query time, reduce the computational and training time, and no need for the offline learning stage. Since the proposed triplet-loss dilated residual network selects the triplets online during the training, it does not require any offline preprocessing and computational steps, and as a result, it has so faster training than other methods.

\section{Conclusion and Future Work}\label{sec5}

The current paper introduced a simple yet efficient method with no added complexity to achieve acceptable accuracy in top-retrieved image tasks. Because the performance of any content-based image retrieval (CBIR) method depends on the representation of the image feature descriptor, the present study uses a deep dilated residual convolutional neural network to capture high-level features with larger receptive fields and produce high-resolution density maps without expanding network complexity. The current research employs the triplet loss function to generate more discriminative representations. To select a set of informative and representative triplets from the training images, an online triplet mining module is used that extracts valid triplets from within each mini-batch. The proposed model obtains promising results in its evaluation of datasets that pose the challenges of changes in appearance, viewpoints, the incomplete view of the object, the lack of sufficiently similar images, and unrelated images. Finally, it is worth noting that the proposed approach currently depends on class labels to choose positive and negative images for each anchor. For future work, the authors intend to create an unsupervised method that can select informative positive and negative with no class label.



\end{document}